\pdfoutput=1

\documentclass[11pt]{article}

\usepackage[final]{acl2025}

\usepackage{times}
\usepackage{latexsym}

\usepackage[T1]{fontenc}

\usepackage[utf8]{inputenc}

\usepackage[utf8]{inputenc}

\usepackage{microtype}

\usepackage{inconsolata}

\usepackage{amsmath}
\usepackage{amsfonts}
\usepackage{graphicx}
\usepackage{booktabs}
\usepackage{multirow}
\usepackage{subcaption}
\usepackage{enumitem}

\DeclareMathOperator*{\argmax}{argmax}

\newcommand{{{\sva}}}{{Speech Vecalign}}
\newcommand{{{\sm}}}{{speech mining}}
\newcommand{{{\gm}}}{{Global Mining}}
\newcommand{{{\lm}}}{{Local Mining}}
\newcommand{{{\mo}}}{{$max\_overlap$}}

\title{{\sva}: an Embedding-based Method for Aligning Parallel Speech Documents}

\author{Chutong Meng\thanks{Work done at Johns Hopkins University.} \\
  George Mason University \\
  \texttt{cmeng2@gmu.edu} \\\And
  Philipp Koehn \\
  Johns Hopkins University \\
  \texttt{phi@jhu.edu} \\}

\begin{document}
\maketitle
\begin{abstract}
We present {\sva}, a parallel speech document alignment method that monotonically aligns speech segment embeddings and does not depend on text transcriptions.
Compared to the baseline method {\gm}~\cite{duquenne-etal-2023-speechmatrix}, a variant of {\sm}, {\sva} produces longer speech-to-speech alignments.
It also demonstrates greater robustness than {\lm}, another {\sm} variant, as it produces less noise.
We applied {\sva} to 3,000 hours of unlabeled parallel English-German~(En-De) speech documents from VoxPopuli, yielding about 1,000 hours of high-quality alignments.
We then trained En-De speech-to-speech translation models on the aligned data.
{\sva} improves the En-to-De and De-to-En performance over {\gm} by 0.37 and 0.18 ASR-BLEU, respectively.
Moreover, our models match or outperform SpeechMatrix model performance, despite using 8 times fewer raw speech documents.\footnote{Data and code are available at \url{https://github.com/mct10/Speech-Vecalign}.}
\end{abstract}

\section{Introduction}
\begin{figure}[t]
    \centering 
    \includegraphics[width=0.85\linewidth]{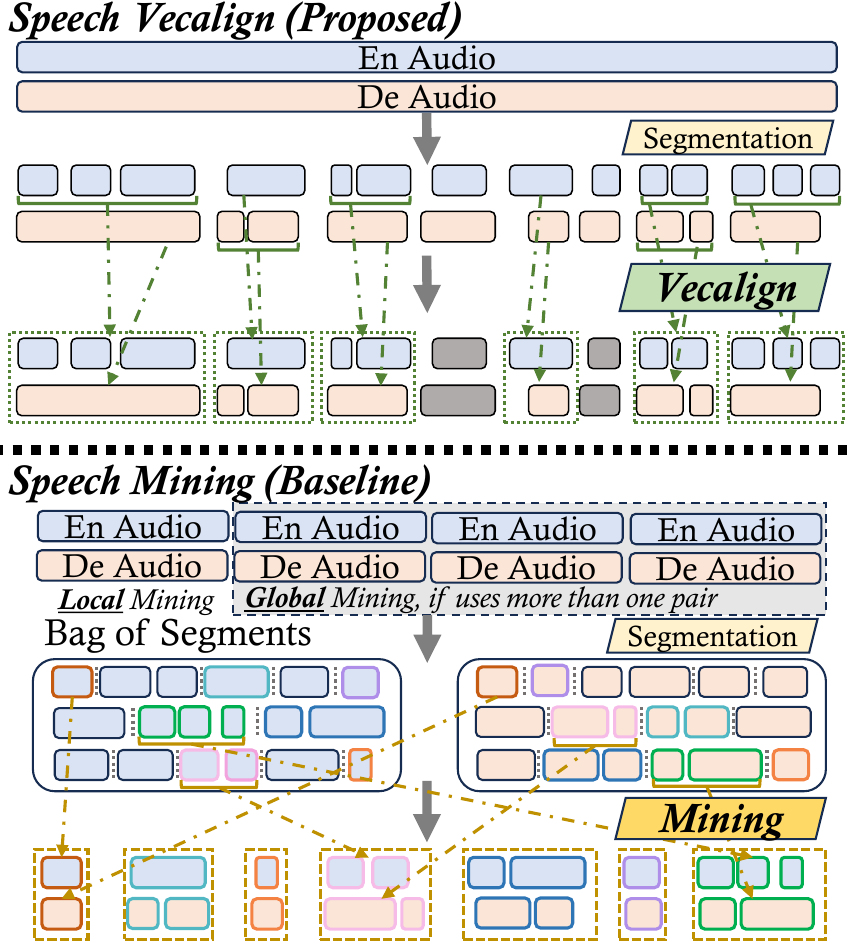}
    \caption{
    A comparison between {\sva} (above) and {\sm} (below). 
    {\sva} aligns each pair of speech documents individually and aligns segments in chronological order, while {\sm} aligns bags of segments and ignores the structure of parallel speech documents.
    }
    \label{fig:fig1}
\vspace{-10pt}
\end{figure}

Speech-to-speech translation (S2ST) is the task of translating speech in one language into speech in another language. 
Conventional S2ST systems concatenate automatic speech recognition (ASR), machine translation (MT), and text-to-speech (TTS) models \cite{lavie-1997-janus, Nakamura-2006-ATR, wahlster-2013-verbmobil}.
These components can be trained individually with datasets for the different components.
Direct S2ST models, which translate source speech into target spectrograms or discrete units with a single architecture, have been recently proposed to alleviate error propagation and to reduce inference latency \cite{jia-2019-direct, lee-etal-2022-direct}. 
Despite the advantages, performance of direct models is limited by the amount of speech-to-speech aligned data, which is much more scarce than the data used for components of cascaded systems.

There have been efforts to automatically curate alignments from multilingual \textit{speech document}s.
In this paper, we define a \textit{speech document} as a file containing more than one utterance and typically comprising several paragraphs, analogous to a \textit{text document}.
VoxPopuli~\cite{wang-etal-2021-voxpopuli} is one such corpus containing a large number of \textit{parallel} speech documents, which are pairs of documents that have the same content but differ in language.

Speech-to-speech alignment methods align short speech clips called \textit{segment}s, and can be either transcription-based or transcription-free.
When transcriptions are available, segments in parallel speech documents can be aligned through speech-to-text and text-to-text alignments.
Inspired by text mining~\cite{schwenk-etal-2021-ccmatrix}, {\sm}~\cite{Duquenne-2021-Multimodal, duquenne-etal-2023-speechmatrix} was proposed as a transcription-free method that aligns speech segments by finding segment pairs with the highest embedding similarity.
It scales well as it does not rely on the availability of text transcriptions.
When {\sm} is applied to a large amount of speech documents, as in all previous work, it is referred to as \textbf{{\gm}}.
Another variant, \textbf{{\lm}}, which applies {\sm} to a single pair of parallel speech documents, has not been well explored.
As we formally define in Section~\ref{sec:speech_mining}, both {\gm} and {\lm} treat documents as bags of unordered segments.

Since {\sm} methods do not leverage the document pair structure, we wonder, \textbf{can we obtain better alignments by aligning speech segments within document pairs and preserving their time order?}
This allows us to utilize the extra knowledge that (1) segments within parallel document pairs are likely to be translations of each other, and (2) segment pairs right next to already aligned pairs are also likely to be aligned.
We draw inspiration from parallel \textit{text} document alignment methods, which have been popular to create sentence-aligned bitext for training MT systems.
Unlike mining, they align sentences for each document pair while maintaining the sentence order.
Our work is based on the text alignment method Vecalign~\cite{thompson-koehn-2019-vecalign}, which aligns parallel sentences by applying fast dynamic time warping~\cite{salvador-2007-toward} to sentence embeddings.
With the advances of extending sentence embeddings to the speech modality~\cite{Duquenne-2021-Multimodal}, we can readily apply Vecalign to parallel speech documents.

In this paper, we introduce {\sva}, a method that aligns parallel speech documents using speech segment embeddings.
Instead of mining from bags of segments, our method aligns individual document pairs and maintains the chronological order of segments, as illustrated in Figure~\ref{fig:fig1}.
Additional preprocessing and postprocessing strategies are applied to improve alignment quality.
We compare {\sva} with {\lm} and {\gm} and show that {\sva} produces higher-quality alignments.
We further provide extensive analysis for all three methods, which could be useful for future research.

\begin{figure*}
    \centering
    \includegraphics[width=0.85\linewidth]{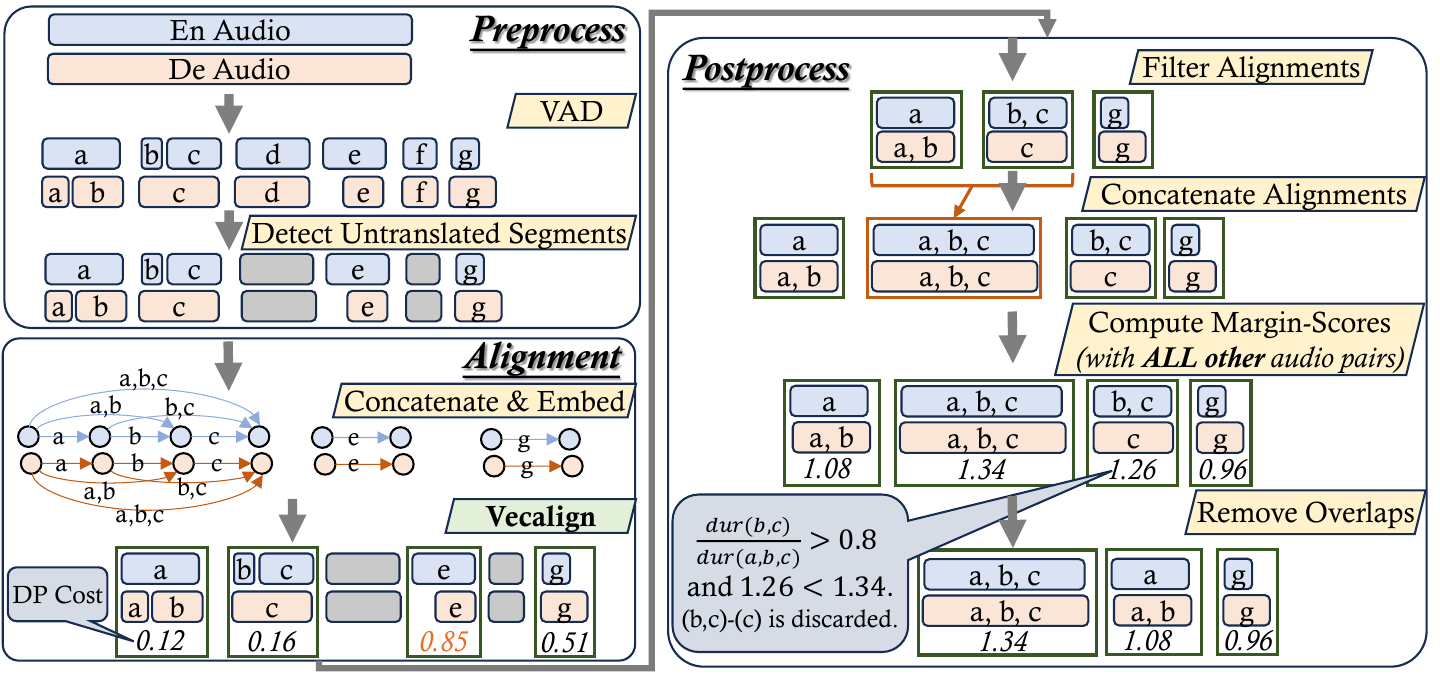}
    \caption{An illustration of the complete {\sva} pipeline using a simple example.
    Each pair of speech documents need to go through 3 steps: Speech Preprocessing (Section \ref{sec:preprocessing}), Segment Alignment (Section \ref{sec:alignment}), and Alignment Postprocessing (Section \ref{sec:postprocessing}).}
    \label{fig:illustration}
\vspace{-10pt}
\end{figure*}

\section{Speech Mining Overview}
\label{sec:speech_mining}
We formally describe the {\sm} methods in this section.
Other related work is in Appendix~\ref{app:related_work}.
 
Speech Mining, first proposed by \citet{Duquenne-2021-Multimodal}, encodes speech segments into language- and modality-agnostic fixed-size embeddings, and then uses margin-based similarity search \cite{artetxe-schwenk-2019-margin} to find the closest embedding pairs.
Depending on the search scope, it can be categorized as {\gm} or {\lm}.

\noindent\textbf{Raw data}. The input data is a list of speech documents $X=[X_1, X_2, \ldots, X_n]$ in the source language and a list $Y=[Y_1, Y_2, \ldots, Y_m]$ in the target language, where $n$ and $m$ are the numbers of documents.
Each document can contain between a few seconds to a few hours of speech.

\noindent\textbf{Speech segmentation}. Voice activity detection~(VAD) is applied to each document to obtain short segments, typically lasting a few seconds.
For instance, $X_i$ is segmented into $X_i=[x_1^i, x_2^i, \ldots, x_{n_i}^i]$,  where ${n_i}$ is the number of segments.
To have segments at different granularities, consecutive segments are progressively concatenated.
$X_i$ becomes $\tilde{X_i}=\left[\tilde{x}_1^i, \tilde{x}_2^i, \ldots, \tilde{x}_{\tilde{n}_i}^i\right]$, where $\tilde{x}$ denotes a concatenated segment and $\tilde{n_i}$ denotes the number of resulting segments.
The same process applies to $Y_j$, producing $\tilde{Y_j}$.

\noindent\textbf{Speech segment embedding}.
Each segment is encoded into a fixed-size embedding using an embedding model.
The segment embeddings for $\tilde{X}_i$ are represented as $E_{\tilde{X}_i}=\left[e^{\tilde{X}_i}_1, e^{\tilde{X}_i}_2, \ldots, e^{\tilde{X}_i}_{\tilde{n}_i}\right]$.
Similarly, the segments in $\tilde{Y_j}$ are encoded as $E_{\tilde{Y_j}}$.

\noindent\textbf{Bag of embeddings}.
In \textit{{\gm}}, embeddings are grouped by \textbf{language}.
We define $G_{X}=\left\{E_{\tilde{X}_1}, E_{\tilde{X}_2}, \ldots, E_{\tilde{X}_n}\right\}$ and $G_{Y}=\left\{E_{\tilde{Y}_1}, E_{\tilde{Y}_2}, \ldots, E_{\tilde{Y}_m}\right\}$, where $G_{X}$ collects all segment embeddings in the source language and $G_{Y}$ collects those in the target language.
In \textit{{\lm}}, embeddings are grouped by \textbf{document pairs}.
Suppose there are $s$ parallel documents, with $X_i$ paired with $Y_i$ for $1\leq i \leq s$.
Documents without a parallel one are ignored.
In this case, $E_{\tilde{X}_i}$ and $E_{\tilde{Y_j}}$ are bags of embeddings themselves.

\noindent\textbf{Embedding alignment}.
Speech mining is performed by finding the most similar embedding pairs between two bags of segment embeddings.
The margin-based similarity, or margin-score, between any two embeddings $a$ and $b$ is computed as
\begin{equation}
\label{eq:margin_score}
\begin{split}
&\text{sim}(a,b)=\\
&\frac{\text{cos}(a,b)}{\sum_{z \in \text{NN}_k(a)}\frac{\text{cos}(a,z)}{2k} + \sum_{z \in \text{NN}_k(b)}\frac{\text{cos}(b,z)}{2k}}
\end{split}
\end{equation}
where $a$ and $b$ are in different languages and $\text{NN}_k(a)$ denotes $k$ nearest neighbors of $a$ in the other language.
The denominator combats the hubness problem.
A higher margin-score indicates better quality.
Then, the mining function for embedding $a$ from a bag of embeddings $B$ is
\begin{align}
\text{mine}(a, B)=\argmax_{b \in B}\ \text{sim}(a, b)
\end{align}
More generally, given two bags of embeddings $U=\{u_1, u_2, \ldots, u_{l_u}\}$ and $V=\{v_1, v_2, \ldots, v_{l_v}\}$, where $l_u$ and $l_v$ are number of embeddings, the collection of all {\sm} alignments is
\begin{align}
\begin{split}
&\text{align}(U, V) \\
&=\left\{(u_1, \text{mine}(u_1, V)), \ldots, (u_{l_u}, \text{mine}(u_{l_u}, V))\right\} \\
&\bigcup \left\{(\text{mine}(v_1, U), v_1), \ldots, (\text{mine}(v_{l_v}, U), v_{l_v})\right\}
\end{split}
\end{align}
Finally, we define {\lm} and {\gm} as
\begin{align}
&\text{Global-Mine}(X, Y)=\text{align}(G_X, G_Y) \\
&\text{Local-Mine}(X,Y)=\bigcup_{i=1}^{s}\text{align}(E_{\tilde{X_i}}, E_{\tilde{Y_i}})
\end{align}

\section{Proposed Method: {\sva}}
\label{sec:method}

The {\sva} pipeline consists of three steps: 
speech preprocessing (Section \ref{sec:preprocessing}), segment alignment with Vecalign (Section \ref{sec:alignment}), and alignment postprocessing (Section \ref{sec:postprocessing}).
An illustration of our method is shown in Figure \ref{fig:illustration}.

\subsection{Speech Preprocessing}
\label{sec:preprocessing}
Speech preprocessing consists of document segmentation and detection of identical untranslated segments.

\noindent \textbf{Segmentation}.  Same as {\sm}, we first segment each speech document by VAD. 
We apply Silero VAD~\cite{Silero-2021-silero}.

\noindent \textbf{Detection of identical untranslated segments}.
As mentioned by \citet{wang-etal-2021-voxpopuli}, some source and target segments contain identical untranslated content due to recording issues.
We introduce this additional step to detect such pairs of segments \textit{prior to} applying the alignment algorithms, in order to make sure they are not aligned.

To find potentially identical untranslated segment pairs, we use a \textit{location heuristic} that they tend to locate in roughly the same position within the source and target documents.
For instance, within each pair of parallel documents, for a source segment $x_a^i$ spanning timestamp $s_{x_a}^i$ to $e_{x_a}^i$, we search for a target segment $y_b^i$ whose midpoint $\frac{s_{y_b}^i+e_{y_b}^i}{2}$ is closest to $\frac{s_{x_a}^i+e_{x_a}^i}{2}$, midpoint of $x_a^i$, since the untranslated target segment is very likely to have a similar time span ($s_{y_b}^i\approx s_{x_a}^i$, $e_{y_b}^i\approx e_{x_a}^i$).

If the two segments have both similar durations and filterbank features, we classify them as identical.
For durations, we compute the time difference.
For filterbank feature, we compute Equation~\ref{eq:fbank_dist}:
\begin{align}
\label{eq:fbank_dist}
\begin{split}
&\text{sim}(\mathbf{A}, \mathbf{B}) \\
&=\min_{i \in \{1, \ldots, T_2-T_1+1\}}\left\{ \frac{1}{NT} \|\mathbf{A}-\mathbf{B_{:, i:i+T_1-1}}\|_F^2 \right\}     
\end{split}
\end{align}
$\mathbf{A} \in \mathbb{R}^{N \times T_1}$ and $\mathbf{B} \in \mathbb{R}^{N \times T_2}$ are  filterbank features\footnote{We use torchaudio.compliance.kaldi.fbank.}  with $N=80$ mel-frequency bins.
$T_1$ and $T_2$ are numbers of frames.
Without loss of generality, we assume $T_1 \leq T_2$.
$\mathbf{B_{:, i:i+T_1-1}} \in \mathbb{R}^{N \times T_1}$ denotes a slice of $\mathbf{B}$ from frame $i$ to frame $i+T_1-1$.
$\|\cdot\|_F$ denotes the Frobenius norm, and $\frac{1}{NT} \|\mathbf{A}-\mathbf{B_{:, i:i+T_1-1}}\|_F^2$ is the mean squared error between $\mathbf{A}$ and a slice of $\mathbf{B}$.
We check if $\mathbf{A}$ could be identical to any slice of $\mathbf{B}$, tolerating the leading and/or trailing noise or silence frames in $\mathbf{B}$.

We empirically determine 0.1 second and $\text{sim}(\mathbf{A}, \mathbf{B})=5.0$ as the thresholds for duration and filterbank similarities.
Note that we cannot depend on speech embeddings for detection, as speech encoders are multilingual and their embeddings are language-agnostic.

\subsection{Speech Segment Alignment}
\label{sec:alignment}
We perform segment alignment based on the similarity between speech segment embeddings.
Unlike {\sm}, which solely relies on similarity scores, we use a dynamic programming~(DP) algorithm to align segments in chronological order.

\noindent \textbf{Segment concatenation}.
Speech segments do not necessarily correspond to complete sentences. 
Same as {\sm}, we first progressively concatenate each segment with the subsequent ones.
Each concatenated segment can contain up to 5 original segments and span a maximum of 20 seconds.

\noindent \textbf{Obtaining segment embeddings}.
After concatenations, we obtain speech segment embeddings using SpeechLASER models~\cite{Duquenne-2021-Multimodal,duquenne-etal-2023-speechmatrix}.
Identical untranslated segments detected in Section \ref{sec:preprocessing}, along with all concatenated segments that include them, are skipped and replaced with $0$-valued vectors.

\noindent \textbf{Applying Vecalign to embeddings}.
We follow~\citet{thompson-koehn-2019-vecalign} to define the cost of aligning two segment embeddings, which serves as the cost function for DP: 
\begin{equation}
\label{eq:dp_cost}
\begin{split}
c(x,y) =
\frac{(1- \text{cos}(x,y)) \text{nSegs}(x) \text{nSegs}(y)}{\sum_{s=1}^S \frac{1-\text{cos}(x, y_s)}{2S} + \sum_{s=1}^S \frac{1-\text{cos}(x_s, y)}{2S}}
\end{split}
\end{equation}
$x$ and $y$ are segment embeddings.
$\text{cos}(\cdot,\cdot)$ computes the cosine similarity.
$x_s$ and $y_s$ are uniformly sampled source and target embeddings, and $S$ is the sample size.
$\text{nSegs}(x)$ is used to denote the number of original segments in $x$, which penalizes aligning long concatenations.

The embedding alignment algorithm is recursive DP.
Given a document pair and corresponding embeddings, the algorithm recursively averages every two consecutive embeddings, halving the sequence length until it reaches a small threshold.
At the bottom level, standard DP is applied to obtain an initial alignment.
Subsequently, at each recursion level bottom-up, DP refines the alignment by searching within a small window around the alignment path from the previous level.
By constraining the search space and reducing the sequence length at each level, the algorithm achieves a linear time and space complexity.
The recursive DP algorithm runs on CPU and takes a few seconds on average per document pair.
We direct the readers to \citet{thompson-koehn-2019-vecalign} for a complete description.

Because of DP, the resultant alignments strictly follow chronological order.
We use $x_{a:b}^i$ to denote the concatenation of consecutive segments $x_{a}^i$ through  $x_{b}^i$.
For any two alignments $(x_{a_s:a_e}^i, y_{b_s:b_e}^j)$ and $(x_{c_s:c_e}^i, y_{d_s:d_e}^k)$, {\sva} guarantees that $i=j=k$ and that either $a_e<c_s,b_e<d_s$ or $a_s>c_e,b_s>d_e$.
In contrast, {\lm} ensures $i=j=k$ but has no constraints on $a,b,c,d$, while
{\gm} makes no guarantees at all.

\subsection{Alignment Postprocessing}
\label{sec:postprocessing}
The goal of postprocessing is to clean the raw alignments and construct alignments with longer durations to improve S2ST models.

\noindent \textbf{Removing low-quality alignments}.
First, we remove unaligned segments and high-cost alignments.
The unaligned segments are due to deletions in the DP algorithm.
Identical untranslated segments detected in Section \ref{sec:preprocessing} may fall into either category due to their 0-valued vectors.

\noindent \textbf{Detection of identical untranslated segments, again}.
Occasionally, the location heuristic in Section \ref{sec:preprocessing} may fail, resulting in a small number of low-cost alignments with identical untranslated source and target segments.
Searching is not needed at this step, as we already have the alignments.
We apply Equation \ref{eq:fbank_dist} to remaining alignments, where $\mathbf{A}$ and $\mathbf{B}$ are source and target segments in each alignment.
We use the same thresholds in Section \ref{sec:preprocessing} to remove alignments.

\noindent \textbf{Alignment concatenation}.
Another issue is that the raw alignments are too short: the average duration is 4.25 seconds, with 66\% shorter than 5 seconds.
To cover more context, we progressively concatenate each alignment with the subsequent ones.
This can be easily done as alignments are in chronological order.
Each concatenated alignment can contain up to 3 original alignments and span up to 20 seconds.

\noindent \textbf{Global margin-scores computation}.
The raw alignments only have alignment costs as a quality indicator, which are computed \textit{within} each document pair.
To assess alignment quality \textit{across} document pairs, we train FAISS~\cite{johnson-etal-2019-billion} indexes and compute margin-scores~\cite{artetxe-schwenk-2019-margin} using Equation~\ref{eq:margin_score} for \textit{all} obtained alignments, following the common strategy in MT dataset curation~\cite{sloto-etal-2023-findings}.

\noindent \textbf{Removing highly-overlapped alignments}.
Finally, we remove alignments that have too much overlap with others, following \citet{duquenne-etal-2023-speechmatrix}.
For any two consecutive alignments, we compute the ratio of the overlapped source duration to the maximum duration of the two source segments.
If the ratio exceeds a threshold, we discard the one with a lower margin-score.
We train S2ST models with multiple threshold values to determine the best one.
Our experiments in Appendix~\ref{app:max_overlap} suggest that 0.4 work best for {\gm} and 0.8 work best for {\lm} and {\sva}.

\section{Experiments \& Results}
We apply {\sva}, {\gm}, and {\lm} to the same raw data and train S2ST models on each type of alignments, providing a fair comparison.

\subsection{Training Data}
\label{sec:train_data}

\textbf{Data source.}
We use the unlabeled, unsegmented English and German plenary session recordings from VoxPopuli v1~\cite{wang-etal-2021-voxpopuli} as raw data.
VoxPopuli contains European Parliament plenary session recordings in each of the 23 European Union languages, paired with spoken interpretations into the other languages.
The document names are formatted as \verb|${session_id}_${language}.ogg|, and paired documents have the same \verb|${session_id}|.
To avoid overlapping with the test set (Section \ref{sec:eval_data}), we only choose sessions from year 2013 to 2020.
We also exclude sessions in the development set~(Section \ref{sec:eval_data}).
For En-to-De, the remaining data has 4,880 documents totaling about 3,000 hours for each language.
For De-to-En, there are 5,782 documents totaling 3,400 hours per language.
The difference is due to the different dev and test sets.
All documents are in pairs, allowing all methods to have exactly the same raw data.

\noindent\textbf{{\sva}.}
We apply {\sva} to each pair of speech documents and obtain alignments sorted by margin-scores.
Training data is chosen in descending order of margin-scores.
We train models on different data sizes and report the best results in Section \ref{sec:results}.
More details on data size optimization can be found in Appendix \ref{app:train_amount}.

\noindent\textbf{Speech mining baselines.}
We apply {\gm} and {\lm} to the same raw data and embeddings as {\sva}.
The implementation is based on \texttt{stopes}\footnote{\url{https://github.com/facebookresearch/stopes}}~\cite{andrews-etal-2022-stopes}.
After mining, we apply the same postprocessing strategies in Section~\ref{sec:postprocessing}, except for alignment concatenation which is not applicable.
Training data is chosen in descending order of margin-scores and details on data size optimization can be found in Appendix~\ref{app:train_amount}.

\subsection{Evaluation Data}
\label{sec:eval_data}
\noindent\textbf{Development set.}
Following \citet{duquenne-etal-2023-speechmatrix}, we choose 1000 samples from the highest scored sessions from the Voxpopuli S2ST dataset.
Additionally, we avoid choosing sessions that occur on the same dates as the test set.

\noindent\textbf{Test set.}
We use the Europarl-ST (EPST) test set~\cite{Iranzo-Sánchez-etal-2020-europarl-st} as an in-domain test set to evaluate the S2ST models. 
EPST is a multilingual S2TT dataset built on European Parliament debates from year 2008 to 2012.
We also adopt FLEURS~\cite{Conneau-etal-2023-FLEURS} as an out-of-domain test set.

\subsection{Experiment Setup}
\label{sec:exp_setup}

We train speech-to-unit translation (S2UT) models~\cite{lee-etal-2022-direct} with \texttt{fairseq}\footnote{\url{https://github.com/facebookresearch/fairseq}} \cite{ott-etal-2019-fairseq,wang-etal-2020-fairseq} on each type of alignments.
The S2UT model takes source speech as input and predicts a sequence of target discrete units.
The discrete units are obtained by applying 
a k-means model to the $11^{\text{th}}$ layer features of a HuBERT model \cite{hsu-etal-2021-hubert}.
For English, we use the mHuBERT from \citet{lee-etal-2022-textless}, and for German, we use the Germanic mHuBERT from \citet{duquenne-etal-2023-speechmatrix}.
Consecutive duplicated units are removed.
Our S2UT model architecture follows exactly \citet{duquenne-etal-2023-speechmatrix}.
The architecture details and training hyperparameters are in Appendix \ref{app:s2st}.

\begin{table*}[!t]
    \footnotesize
    \centering
    \begin{tabular}{@{\extracolsep{1.5pt}}cl*{5}c}
        \toprule
         & \multicolumn{2}{c}{\textbf{Training Data}} & \multirow{2}{*}{\textbf{ASR-BLEU}} & \multirow{2}{*}{\textbf{ASR-chrF2++}} & \multicolumn{2}{c}{\textbf{BLASER 2.0}} \\
         \cmidrule{2-3} \cmidrule{6-7}
         & Alignment Method & \# Hours & & & w/ text ref & w/o text ref \\
         \midrule
         & \multicolumn{1}{l}{\colorbox{lightgray}{\textbf{English-to-German}}} \\
         \multirow{2}{*}{\textit{\textbf{State-of-the-art}}} & SpeechMatrix$^\dagger$ & 1451 & 10.1  & - & - & - \\
         & SpeechMatrix$^\ddagger$ & 1451 & 11.27  & 39.98 & 3.52 & 3.86 \\
         \midrule
         \multirow{2}{*}{\textit{\textbf{Baseline}}} & {\lm} & 1500 & \underline{\textbf{12.91}} & 44.08 & 3.70 & 4.02 \\
         & {\gm} & 1000 & 12.21 & 42.65 & 3.65 & 3.97 \\
         \midrule
         \textit{\textbf{Our Method}} & {\sva} & 750 & 12.58 & \underline{\textbf{44.34}} & \underline{\textbf{3.73}} & \underline{\textbf{4.05}} \\
         \midrule
         & \multicolumn{2}{l}{p-value w.r.t {\lm}} & 0.1069 & 0.0999 & 0.0050\textsuperscript{*} & 0.0020\textsuperscript{*} \\
         & \multicolumn{2}{l}{p-value w.r.t {\gm}} & 0.0769 & 0.0010\textsuperscript{*} & 0.0010\textsuperscript{*} & 0.0010\textsuperscript{*} \\
         \midrule
         \midrule
         & \multicolumn{1}{l}{\colorbox{lightgray}{\textbf{German-to-English}}} \\
         \multirow{2}{*}{\textit{\textbf{State-of-the-art}}} & SpeechMatrix$^\dagger$  & 1456 & 16.3  & - & - & -\\
         & SpeechMatrix$^\ddagger$  & 1456 & \underline{16.62}  & \underline{43.77} & \underline{3.81} & \underline{4.11} \\
         \midrule
         \multirow{2}{*}{\textit{\textbf{Baseline}}} & {\lm} & 1250  & 15.64  & 42.85 & 3.70 &  4.02 \\
         & {\gm} & 750 & 15.96 &  43.14 & 3.74 & 4.04 \\
         \midrule
         \textit{\textbf{Our Method}} & {\sva} & 1000 & \textbf{16.14} & \textbf{43.71} & \textbf{3.76} &  \textbf{4.07} \\         
         \midrule
         & \multicolumn{2}{l}{p-value w.r.t {\lm}} & 0.0030\textsuperscript{*} & 0.0010\textsuperscript{*} & 0.0010\textsuperscript{*} & 0.0010\textsuperscript{*} \\
         & \multicolumn{2}{l}{p-value w.r.t {\gm}} & 0.1449 & 0.0030\textsuperscript{*} & 0.0030\textsuperscript{*} & 0.0010\textsuperscript{*} \\
         \bottomrule
    \end{tabular}
    \caption{Results for En-to-De and De-to-En on EPST test sets. \textbf{Bold} means better than {\sm} baselines. \underline{Underline} means the best overall.
    $^\dagger$Results from \citet{duquenne-etal-2023-speechmatrix}.
    $^\ddagger$Models trained by ourselves.
    \textsuperscript{*}p-value $<0.05$.
    Results show that {\sva} models perform better than baselines under almost all metrics in both directions.
    }
    \label{tab:main_result}
\vspace{-10pt}
\end{table*}

\subsection{Evaluation Metrics}
With the discrete units generated by S2UT models, we resynthesize speech using pretrained unit-based HiFi-GAN vocoders \cite{polyak-etal-2021-speech} from \citet{duquenne-etal-2023-speechmatrix}.
We then evaluate the resynthesized speech using both transcription-based and transcription-free methods.

For the transcription-based method, we transcribe the speech output using the same ASR models as \citet{duquenne-etal-2023-speechmatrix}.
We evaluate the transcriptions using SacreBLEU\footnote{\url{https://github.com/mjpost/sacrebleu}} \cite{post-2018-call} to compute BLEU\footnote{Signature: nrefs:1 + case:mixed + eff:no 
+ tok:13a + smooth:exp + version:2.2.0} and chrF2++\footnote{Signature: nrefs:1 + case:mixed + eff:yes + nc:6 + nw:2 + space:no + version:2.2.0} scores.
We apply the significance test using paired bootstrap resampling~\cite{koehn-2004-statistical} with 1000 bootstrap resamples.

We also adopt BLASER 2.0 \cite{dale-costa-jussa-2024-blaser} to directly evaluate speech output.
We compute the referenced score using \verb|blaser-2.0-ref|\footnote{\url{https://huggingface.co/facebook/blaser-2.0-ref}} for input and output speech, as well as the text reference.
We compute the reference-free score using \verb|blaser-2.0-qe|\footnote{\url{https://huggingface.co/facebook/blaser-2.0-qe}} for input and output speech only.

\subsection{Main Results}
\label{sec:results}
As mentioned in Section \ref{sec:train_data}, we train models on data of various sizes.
Table \ref{tab:main_result} presents the best En-to-De and De-to-En results on the EPST test set, along with the corresponding data sizes.
Additional results on the FLEURS test set are in Appendix \ref{app:fleurs_result}.

Intriguingly, for both directions, {\sva} and {\sm} models are competitive with or outperform SpeechMatrix~\cite{duquenne-etal-2023-speechmatrix} models, despite the latter being mined from about 24k hours of speech per language, \textit{8 times more} than our raw data.\footnote{We do not aim for state-of-the-art performance. Our results are not directly comparable to SpeechMatrix. We report SpeechMatrix results only to show the performance gap.}
For En-to-De, our {\gm} and {\sva} models achieve improvements of 0.94 and 1.31 BLEU, respectively.
Our {\lm} model achieves even 1.64 BLEU improvement.
We suspect that SpeechMatrix has not removed identical untranslated segments prior to and after mining, which significantly hurts model performance.
Further discussion is in Section \ref{sec:identical}.

While {\lm} has not been previously explored, our results suggest that it is a potentially useful method.
{\lm} achieves the highest BLEU score in En-to-De, and only slightly underperforms {\gm} in De-to-En, indicating that constraining the mining scope to document pairs does not necessarily have a negative impact on alignment quality.
Yet we note that {\lm} requires more training data to achieve its optimal performance, as shown in Appendix \ref{app:train_amount}.

Our {\sva} models outperform both {\sm} models in both directions.
For En-to-De, the {\sva} model achieves 12.58 BLEU, comparable with our strong {\gm} and {\lm} baselines. 
In terms of chrF2++, it surpasses {\gm} and {\lm} by 1.69 and 0.26, respectively.
It also significantly improves their referenced BLASER 2.0 by 0.08 and 0.03.
For De-to-En, {\sva} and {\gm} models achieve comparable BLEU (16.14 vs. 15.96), but {\sva} surpasses {\gm} by 0.57 in chrF2++.
{\sva} significantly outperforms {\lm} under all metrics.
These results demonstrate that {\sva} produces higher-quality alignments than both {\sm} baselines.

\section{Analysis}
We analyze properties of {\sm} methods and compare them with {\sva}.
Although we show that {\sm} methods produce alignments similar to those of {\sva}, the latter offers advantages of producing longer and less noisy alignments.

\subsection{Speech Mining Mostly Locally Aligns Segments in Time Order}
\label{sec:sm_aligns_pairs}
First, we show that {\gm} mostly \textbf{locally} aligns speech documents.
While {\gm} searches for the best matching segment pairs among roughly 10 million segments, one might expect its alignments to cover the spread of the entire dataset.
On the contrary, we find that {\gm} alignments are concentrated within document pairs, each typically containing hundreds to thousands of segments.

\begin{figure}[h]
    \centering
    \includegraphics[width=0.9\linewidth]{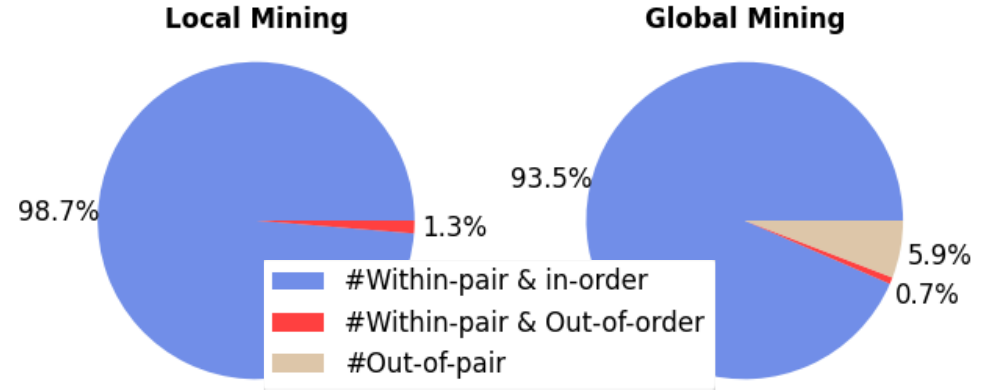}
    \caption{For both 1000-hour {\sm} datasets, we compute percentages of En-De alignments that come from different document pairs, and those are within paired documents but out-of-order.
    {\gm} has $5.9\%$ out-of-pair alignments, and both methods have only around $1\%$ out-of-order alignments.
    }
    \label{fig:non_local_alignments}
\vspace{-10pt}
\end{figure}

To quantify this, we examine the 1000-hour {\gm} data and count alignments whose source and target segments come from \textit{different} document pairs.
As shown in Figure~\ref{fig:non_local_alignments}, fewer than 6\% fall into this category, while the majority~($>93\%$) are within paired documents.

Second, we analyze the time order of alignments produced by both {\sm} methods.
Borrowing the notation from Section~\ref{sec:alignment}, we define two pairs of alignments to be \textit{in-order} if either $a_e<c_s,b_e<d_s$ or $a_s>c_e,b_s>d_e$; otherwise, they are \textit{out-of-order}.
Figure \ref{fig:non_local_alignments} shows that only around 1\% alignments are out-of-order for both {\sm} methods.

Observations above indicate that {\sm} alignments are mostly within paired documents and preserve time order.
We hypothesize that speech-to-speech alignments are sparse and high-quality ones mostly exist in paired documents.

As a by-product, this property can be leveraged to identify parallel documents.
If {\gm} finds many alignments between two documents, they are likely to be parallel.
It is particularly useful when the pairing metadata is not readily available.

\subsection{Speech Mining Methods Produce Similar Alignments as {\sva}}
\label{sec:sv_similar_to_sm}

Following the observations in Section \ref{sec:sm_aligns_pairs} that {\sm} produces mostly local, in-order alignments, we analyze the similarity between them and {\sva} alignments.

We employ the alignment evaluation method\footnote{\url{https://github.com/thompsonb/vecalign/blob/master/score.py}.} from~\citet{thompson-koehn-2019-vecalign}, which computes precision and recall by comparing system alignments to a reference.
There are two modes: \textit{Strict}, which counts only exact matches as true positives, and \textit{Lax}, which considers an alignment as true positive if both its source and target segment overlap with the reference.
If not true positive, an alignment is false positive.
Recall is computed by swapping the reference and the system alignments.

Without loss of generality, we use {\sva} En-De alignments as the reference, and evaluate {\sm} ones.
We choose 700k highest-scoring alignments from all three methods to ensure a fair comparison.
Table \ref{tab:sv_gm_sim} shows that about 30\% of {\sm} alignments are exactly the same as those of {\sva}, and about 90\% overlap with {\sva} alignments.
This high similarity explains why {\sva} and {\sm} models have similar performance.

\begin{table}[th]
    \centering
    \small
    \begin{tabular}{lcccc}
        \toprule
        \multirow{2}{*}{\textbf{Mode}} & \multicolumn{2}{c}{\textbf{{\gm}}} & \multicolumn{2}{c}{\textbf{{\lm}}} \\
        \cmidrule{2-3} \cmidrule{4-5}
         & Precision & Recall & Precision & Recall \\
        \midrule
        \textit{Strict} & 0.325 & 0.326 & 0.305 & 0.305 \\
        \textit{Lax} & 0.865 & 0.965 & 0.963 & 0.814 \\
        \bottomrule
    \end{tabular}
    \caption{Precision and Recall for {\sm} alignments when {\sva} is used as the reference.
    The high precision and recall in the \textit{Lax} mode indicate the methods produce similar alignments.
    }
    \label{tab:sv_gm_sim}
\vspace{-10pt}
\end{table}

\subsection{{\sva} Produces Longer Alignments}

As {\sm} and {\sva} produce similar alignments, we explore why {\sva} models still perform better.
A key advantage of {\sva} is that it first produces fine-grained alignments and then constructs alignments with different amounts of context, thanks to the alignment concatenation strategy.
Speech mining methods, on the other hand, solely depend on margin-scores and tend to favor shorter alignments.

\begin{figure}[h]
    \centering
    \includegraphics[width=\linewidth]{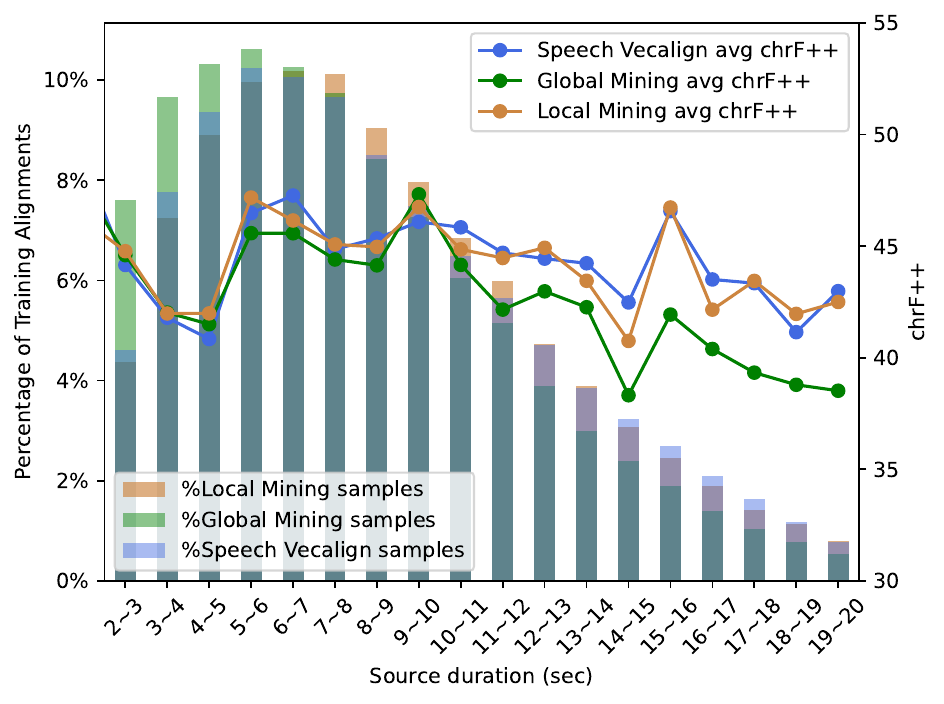}
    \caption{Average sentence-level chrF2++ on En-to-De test set and the corresponding portion of training samples with respect to duration ranges.
    The {\sva} model consistently performs better on longer inputs.
    }
    \label{fig:chrf_and_dur}
\vspace{-10pt}
\end{figure}

With the best En-to-De models and corresponding data sizes from Section \ref{sec:results}, Figure~\ref{fig:chrf_and_dur} presents the average sentence-level chrF2++ scores on the test set and the percentage of training alignments for different source speech duration ranges.
Notably, {\sva} has a large portion of long training samples: the blue bars are highest for durations longer than 12 seconds.
Specifically, the average source duration of {\sva} is 8.51 seconds, while {\gm} and {\lm} have average durations of 7.50 and 8.53 seconds, respectively.
As a result, the {\sva} model performs better on test samples longer than 10 seconds, while having comparable performance on shorter ones.
This highlights that {\sva} is able to produce longer, context-rich alignments which help to improve S2ST model performance.
Interestingly, {\lm} surpasses the {\gm} model on long inputs, which could be also attributed to its longer training samples.

\begin{figure*}[t]
    \centering
    \begin{minipage}{0.22\linewidth}
        \centering
        \centerline{\includegraphics[width=\linewidth]{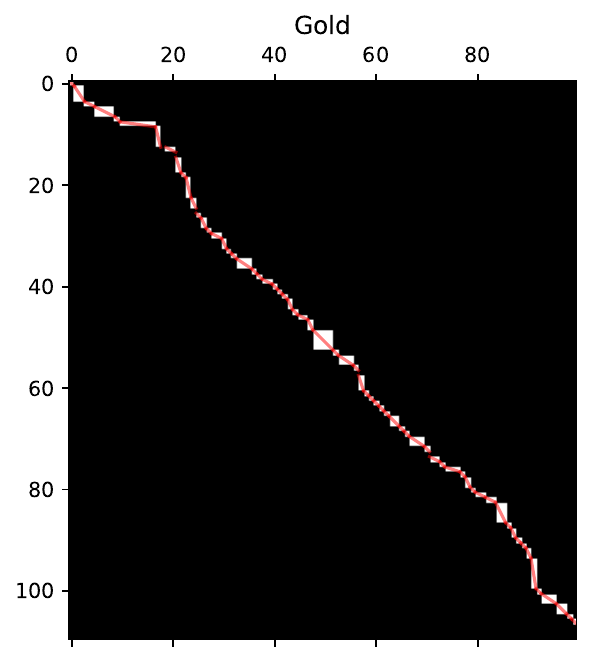}}
        \subcaption{Gold}
    \end{minipage}
    \begin{minipage}{0.22\linewidth}
        \centering
        \centerline{\includegraphics[width=\linewidth]{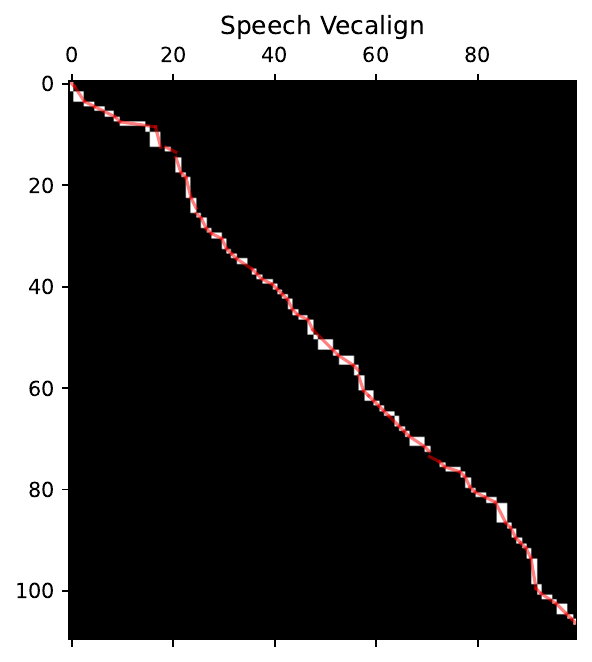}}
        \subcaption{{\sva}}
    \end{minipage}
    \begin{minipage}{0.22\linewidth}
        \centering
        \centerline{\includegraphics[width=\linewidth]{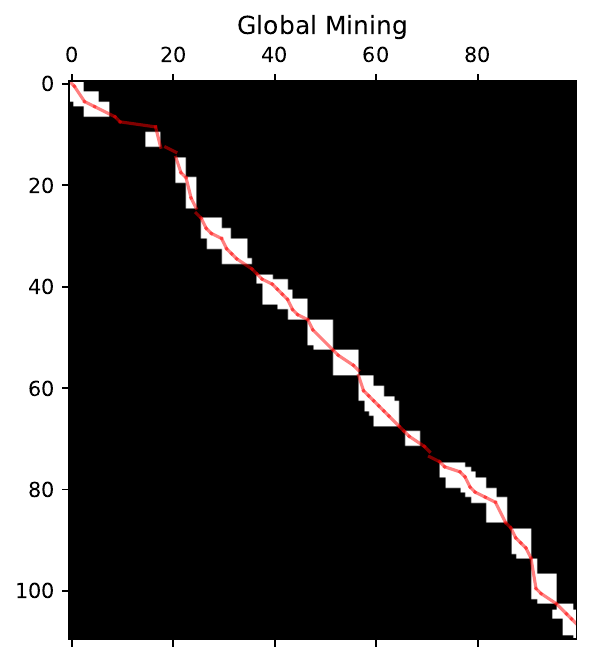}}
        \subcaption{{\gm}}
    \end{minipage}
    \begin{minipage}{0.22\linewidth}
        \centering
        \centerline{\includegraphics[width=\linewidth]{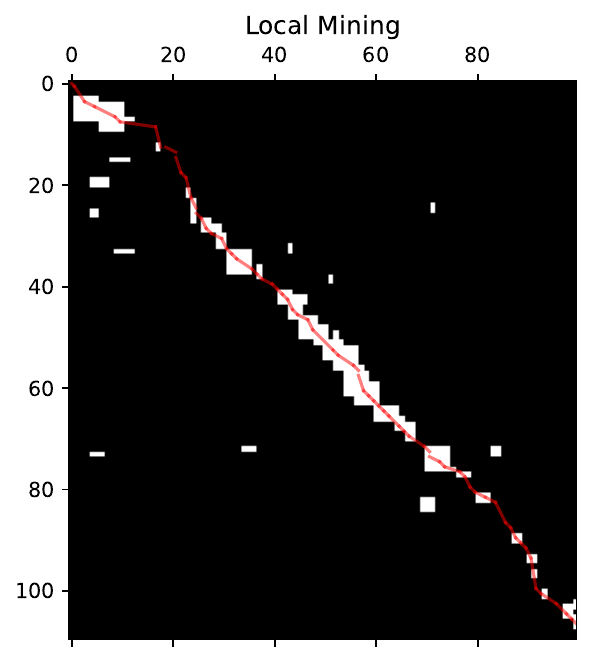}}
        \subcaption{{\lm}}
    \end{minipage}
    \caption{
    Visualizations of gold and 3 system alignments for 20180313-0900-PLENARY-15.
    The red lines indicate the gold alignment.
    Figures from left to right are:
    (a) Gold alignment manually created  by us.
    (b) Raw {\sva} alignments \textit{without} alignment concatenation.
    (c) {\gm}.
    (d) {\lm}.
    {\sva} and {\gm} follow closely with the gold alignment, while {\lm} produces more noise.
    }
    \label{fig:local_mining}
\vspace{-10pt}
\end{figure*}

\subsection{{\sva} Produces Less Noise than {\lm}}
\label{sec:local_mining}

We visualize alignments produced by different methods for the same document pair, which is about 10 minutes long and contains around 200 segments.
For reference, we manually created a gold segment-level alignment, with detailed procedure in Appendix \ref{app:gold_alignment}.
We illustrate the best 80 alignments for each of the {\sm} methods.

As Figure \ref{fig:local_mining} shows, {\sva} produces the most fine-grained alignments and is most similar to the gold reference.
{\gm} also performs well, aligning closely with the groundtruth path, whereas {\lm} produces more noise and misses more alignments along the correct path.
We hypothesize {\lm} has limited number of segments in a single document pair, making nearest neighbors less effective normalizers in the margin-score computation.

\subsection{Removing Identical Untranslated  Segments is Critical}
\label{sec:identical}
As presented in Section \ref{sec:results}, our reproduced speech mining models achieve comparable or even better results than SpeechMatrix models.
By listening to samples of SpeechMatrix alignments, we observed many cases where the source and target segments contained identical untranslated content, which is an issue mentioned in Section \ref{sec:preprocessing}.
Using the method described in Section \ref{sec:preprocessing}, we identified approximately 100k out of 630k alignments with untranslated source and target segments, totaling 181 hours.

To evaluate the impact of untranslated segments, we trained models on the original SpeechMatrix En-De alignments and on a version with untranslated alignments \textit{removed}.
The training data is chosen with a margin-score threshold of 1.09, following the original setup.  
As shown in Table~\ref{tab:clean_speech_matrix}, the cleaned data produces better models, improving BLEU score by \textbf{1.00} for En-to-De and \textbf{0.11} for De-to-En, despite having 13\% less training data.
The smaller gain on De-to-En may be due to most untranslated segments being in English, which have smaller impact on into-English translation.

We also re-produced our alignment pipelines \textit{without} removing identical untranslated segments, referred to as ``noisy" in Table \ref{tab:noisy_3_methods}.
We trained models on 500 hours of this data.
Although these untranslated segments account for less than 1\% of the training data, performance degrades noticeably.

Overall, the experiments highlight that removing untranslated alignments is essential for S2ST training, corroborating \citet{khayrallah-koehn-2018-impact}, who found that the untranslated sentences are most catastrophic in neural machine translation.

\begin{table}[ht]
    \centering
    \small
    \begin{tabular}{ccc}
        \toprule
        \textbf{Dataset} & \textbf{Hours} & \textbf{ASR-BLEU} \\
        \midrule
         {\colorbox{lightgray}{\textbf{English-to-German}}} \\
        SpeechMatrix & 1451 & 11.27 \\ 
        SpeechMatrix cleaned & 1265 & \textbf{12.27} \\ 
        \midrule
         {\colorbox{lightgray}{\textbf{German-to-English}}} \\
         SpeechMatrix & 1456 & 16.62 \\ 
         SpeechMatrix cleaned & 1276 & \textbf{16.73} \\ 
        \bottomrule
    \end{tabular}
    \caption{
    Performance of models trained on SpeechMatrix, before and after removing identical untranslated alignments.
    Results are measured on En-to-De and De-to-En EPST test sets.
    The removal of untranslated segments boosts model performance.
    }
    \label{tab:clean_speech_matrix}
\vspace{-10pt}
\end{table}

\begin{table}[h]
    \centering
    \scriptsize
    \begin{tabular}{lcc}
        \toprule
        \multicolumn{1}{c}{\textbf{Dataset}} & \textbf{\#Noisy/All Aligns.} & \textbf{ASR-BLEU} \\
        \midrule
        {\lm} & - & \textbf{11.18} \\
        {\lm} noisy & 2.66k/236k & 10.76 \\
        \midrule
        {\gm} & - & \textbf{11.54} \\
        {\gm} noisy & 1.46k/254k & 11.27 \\
        \midrule
        {\sva} & - & \textbf{11.78} \\
        {\sva} noisy & 1.48k/222k & 11.69 \\
        \bottomrule
    \end{tabular}
    \caption{Results for En-to-De on EPST test sets.
    ``noisy" means the steps of removing identical untranslated segments are \textbf{not} applied.
    All datasets have 500 hours of training data.
    }
    \label{tab:noisy_3_methods}
\end{table}

\section{Conclusion}
We present {\sva}, a parallel speech document alignment method that aligns speech segment embeddings within document pairs and in chronological order.
We apply {\sva} to parallel English-German VoxPopuli speech documents and conduct S2ST experiments to demonstrate its superiority over two strong {\sm} baselines.
Our analysis reveals that although {\sm} methods primarily align documents locally and in-order, {\gm} falls short of producing long alignments, and {\lm} in particular produces more noise.
For long-term future work, we plan to extend {\sva} to other language pairs or other data sources.
We can also explore aligning speech and text embeddings to construct S2TT datasets.

\section*{Limitations}
\noindent\textbf{Speech features for identical untranslated segment detection could be improved.}
Our current approach uses filterbank features, which are based on power spectrum, to detect identical untranslated segments.
However, filterbank features are likely to fail for segments that have identical content but differ in signal power.
As one of the anonymous reviewers pointed out, cepstral features may be a more robust alternative.

\noindent\textbf{Limited language pair.}
We have only conducted experiments for English and German speech from the VoxPopuli dataset. 
As {\sva} heavily relies on the quality of speech embeddings, the performance is unclear for other language pairs and other domains of speech.

\noindent\textbf{Dependency on parallel speech documents.}
{\sva} requires parallel speech documents, which is often not available. 
We may rely on {\gm} to discover parallel documents, as Section \ref{sec:sm_aligns_pairs} suggests, but doing so will introduce extra computation costs.

\section*{Acknowledgments}
This work was in part supported by a Sony Research Award.
We thank Antonios Anastasopoulos for proofreading an early version of this paper.
We are thankful to anonymous reviewers for their valuable feedback.

\bibliography{anthology,custom}

\appendix

\section{Related Work}
\label{app:related_work}
\textbf{Speech-to-speech translation (S2ST).}
The early S2ST systems consist of cascaded ASR, MT, and TTS models \cite{lavie-1997-janus, Nakamura-2006-ATR, wahlster-2013-verbmobil}.
Direct S2ST models have recently been proposed to alleviate error propagation, support unwritten languages, and improve inference speed.
Translatotron models \cite{jia-2019-direct,jia-2022-translatotron2} are trained with spectrograms as targets, while the S2UT model~\cite{lee-etal-2022-direct} outputs discrete units.
UnitY~\cite{inaguma-etal-2023-unity} and UnitY2 \cite{communication-2023-seamless} are two-pass direct S2ST models that predict both subwords and discrete units with a single model.
Despite advances in architectures, the amount of supervised training data is still insufficient and thus limits model performance.

\noindent\textbf{Bilingual text sentence alignment.}
Text alignment is very related to speech alignment.
Methods apply dynamic programming \cite{bellman-1954-theory} and mainly differ in the design of scoring functions.
Early works \cite{brown-etal-1991-aligning, gale-church-1993-program} are based on sentence lengths.
Later methods incorporate translations in various ways~\cite{moore-2002-fast, varga-2007-parallel,sennrich-volk-2010-mt, gomes-lopes-2016-first}.
Our work is inspired by Vecalign \cite{thompson-koehn-2019-vecalign}, which utilizes margin-based cosine similarities between multilingual sentence embeddings like LASER~\cite{artetxe-schwenk-2019-massively, heffernan-etal-2022-bitext} and LaBSE \cite{feng-etal-2022-language}. 
Vecalign is also more efficient than previous methods.
By applying fast dynamic time warping \cite{salvador-2007-toward}, it has a linear time and space complexity with respect to the number of input sentences.
The recent progress of extending multilingual sentence embeddings to the speech modality \cite{Duquenne-2021-Multimodal,Khurana-etal-2022-samuxlsr, duquenne-etal-2023-speechmatrix, Duquenne-2023-sonar_arxiv} enables us to align speech segments by their speech embeddings using the same algorithm.

\noindent\textbf{S2ST datasets.}
There are two common ways to automatically build an S2ST dataset: (1)~building alignments from multilingual speech data; (2)~synthesizing speech for text translations from existing speech-to-text translation (S2TT) corpora.
The first line of work has human spoken speech on both source and target sides.
VoxPopuli \cite{wang-etal-2021-voxpopuli} aligns multilingual speech documents based on text transcriptions, yielding 17.3k-hour alignments between 15 source and target languages.
SpeechMatrix \cite{duquenne-etal-2023-speechmatrix} applies {\gm} with SpeechLASER embeddings on VoxPopuli. It obtains alignments for 136 language pairs with an average of 1537 hours per direction.
\citet{communication-2023-seamless} apply {\gm} to web-crawled speech data with SONAR embeddings.
\citet{communication-2023-seamless} also mine a SeamlessAlignExpressive dataset with expressively- and semantically-aligned segment pairs, based on a blend of both semantic and prosodic similarity score \cite{heffernan-etal-2024-aligning}.

The second line of work has synthesized speech on the target side.
Fisher \cite{post-etal-2013-improved} is a Spanish-English S2TT dataset containing about 170 hours of Spanish telephone conversations and English translations which are used to synthesize English speech.
CVSS \cite{jia-etal-2022-cvss} is an S2ST dataset covering utterances from 21 languages to English, obtained by synthesizing the text translations in CoVoST 2 \cite{wang-2021-CoVoST2}.
Besides automatic methods, FLEURS \cite{Conneau-etal-2023-FLEURS} has collected human read speech covering 102 languages. But it contains only about 12 hours per language and is intended for evaluation.

\section{Speech-to-Speech Translation}
\label{app:s2st}

We describe the model architecture in Appendix \ref{app:appendix_arch} and the experiment hyperparameters in Appendix \ref{app:hyper}.

\subsection{S2UT Model Architecture}
\label{app:appendix_arch}
The S2UT model \cite{lee-etal-2022-direct} adopts the Transformer encoder-decoder architecture~\cite{vaswani-etal-2017-attention}.
It has 2 convolutional layers, 12 transformer encoders, and 6 transformer decoders for target unit prediction.
Additionally, 2 transformer decoders for source unit prediction are attached to the $6^{\text{th}}$ encoder layer.
All embedding dimensions are 512, except for the source unit decoder, which has a dimension of 256. 
The forward dimensions are 2048.
The model has a total of 70M trainable parameters.

\subsection{Training Hyperparameters}
\label{app:hyper}
We train the models using a learning rate of 0.0005 with the inverse\_sqrt scheduler.
We use the adam optimizer \cite{kingma2017adammethodstochasticoptimization} with betas $(0.9, 0.98)$.
We apply a dropout rate of 0.3 and a label smoothing weight of 0.2.

Due to limited computing resources, we adopt different training strategies for different purposes.
The 500-hour datasets are used for hyperparameter optimization, and larger datasets are used for reporting main results.
All models are trained for up to 400k steps, with the first 10,000 steps as a warmup stage.
For experiments on a 500-hour dataset, we use a batch size of 320k tokens and apply early-stopping if there is no improvement on the development set for 30 epochs.
These models are trained on 4 NVIDIA GeForce GTX 1080 Ti GPUs for approximately 15 days.
For larger datasets, we increase the batch size to 640k tokens and early-stopping is not applied.
These models are trained on 2 NVIDIA A100-SXM4-80GB GPUs for approximately 15 days.
The best checkpoint is selected based on the development set loss.
All experiments are conducted in fp32, as we found training with fp16 and amp very unstable.

\section{Computation Costs for Alignment}
\noindent\textbf{Segment embedding.}
This is the most time-consuming step.
We use a mixture of NVIDIA GeForce GTX 1080 and 2080 Ti GPUs. 
Embedding about 6,000 hours of speech (3,000 hours for each language) took approximately 1,100 GPU hours.

\noindent\textbf{Alignment.}
{\lm} and {\gm} run on a single GPU.
They take about 2 hours.
{\sva} runs on a single CPU and takes about 2 hours.

\section{Training Data Optimization}

There are two hyperparameters that affect training data: (1) the maximum source duration overlap ratio between alignments, {\mo}, which is mentioned in Section \ref{sec:postprocessing}, and
(2) the data size.

{\mo} controls the trade-off between overlapped durations and data quality.
For instance, a lower {\mo} reduces the overlap but also discards alignments more aggressively.
Overlapped alignments usually have similar margin-scores, so more high-quality alignments are lost.
The data size controls the trade-off between data size and data quality cutoff.
For instance, a larger dataset will have a lower quality cutoff, as alignments are selected in descending order of margin-scores.
In this section, we optimize the combination of {\mo} and data size by training S2UT models on different datasets.
Note that the raw data stays the same. 

\subsection{Optimizing {\mo}}
\label{app:max_overlap}

\begin{figure}[h]
    \centering
    \includegraphics[width=0.95\linewidth]{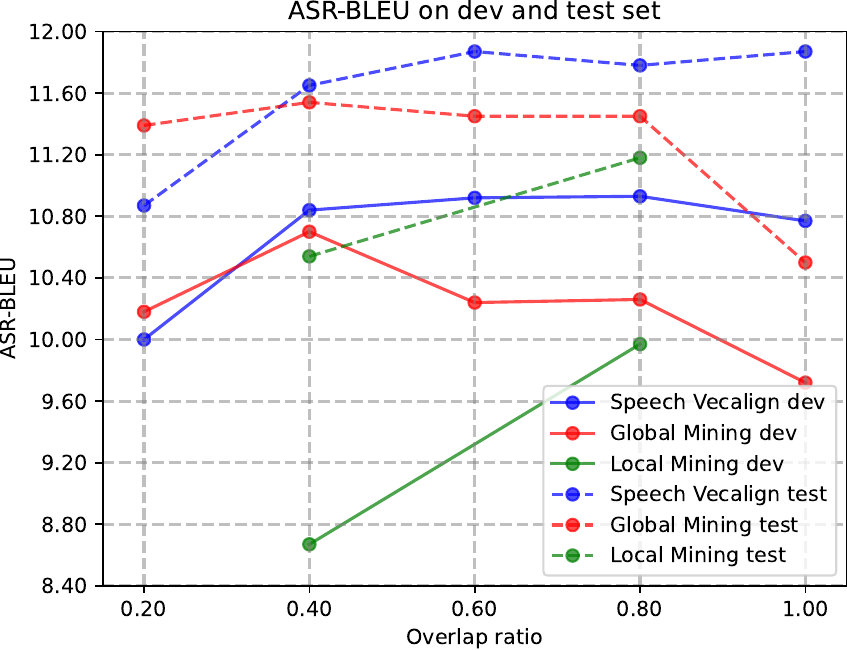}
    \caption{ASR-BLEU on En-to-De EPST dev and test set.
    All models are trained with 500-hour data. Only $max\_overlap$ varies.}
    \label{fig:dev_bleu_overlap_ratio}
\vspace{-10pt}
\end{figure}

We first experiment with different values of {\mo}.
We apply different {\mo} thresholds during the postprocessing stage, and always choose the best 500 hours as the training data.
The optimal value is determined based on development set ASR-BLEU. 
Figure \ref{fig:dev_bleu_overlap_ratio} shows that 0.8 works best for {\sva} and {\lm} and 0.4 works best for {\gm}.
The test set performance is also drawn in Figure \ref{fig:dev_bleu_overlap_ratio}, exhibiting a similar trend.

\subsection{Optimizing Training Data Size}
\label{app:train_amount}
\begin{figure}[h]
    \centering
    \includegraphics[width=0.9\linewidth]{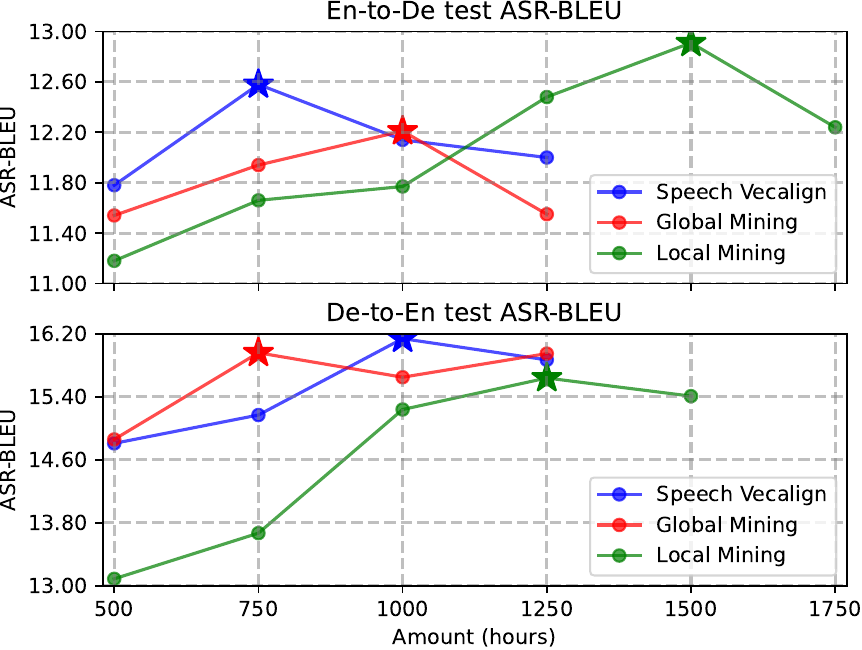}
    \caption{ASR-BLEU on En-to-De (above) and De-to-En (below) EPST test set. Models are trained on different amounts of training data.
    The best {\sva} models outperform the other {\sm} models.
    }
    \label{fig:test_bleu_amount}
\vspace{-10pt}
\end{figure}

Next we optimize the training data size.
We fix {\mo} at 0.4 for {\gm} and 0.8 for {\sva} and {\lm} during postprocessing, only lowering the quality cutoff to include more training data.
The models are trained on different amounts of data until we find the peak performance.
Results are shown in Figure~\ref{fig:test_bleu_amount}.

For En-to-De, the best {\sva} model is trained on the 750-hour dataset, achieving 12.58 BLEU.
It outperforms the best {\gm} model which achieves 12.21 BLEU.
The best {\lm} model achieves 12.91 BLEU.
However, we note that it requires a lot more data than the other two methods to achieve the peak performance.

For De-to-En, the 1000-hour dataset works best for {\sva} while the 750-hour dataset works best for {\gm}.
{\lm} achieves the peak performance when the data size is 1250 hours, still requiring more data than the other methods.
The {\sva} performs better than both the {\gm} and the {\lm} models.

\begin{table*}[!t]
    \footnotesize
    \centering
    \begin{tabular}{@{\extracolsep{1.5pt}}cl*{5}c}
        \toprule
         & \multicolumn{2}{c}{\textbf{Training Data}} & \multirow{2}{*}{\textbf{ASR-BLEU}} & \multirow{2}{*}{\textbf{ASR-chrF2++}} & \multicolumn{2}{c}{\textbf{BLASER 2.0}} \\
         \cmidrule{2-3} \cmidrule{6-7}
         & Alignment Method & \# Hours & & & w/ text ref & w/o text ref \\
         \midrule
         & \multicolumn{1}{l}{\colorbox{lightgray}{\textbf{English-to-German}}} \\
         \multirow{2}{*}{\textit{\textbf{State-of-the-art}}} & SpeechMatrix$^\dagger$ & 1451 & 2.7  & - & - & - \\
         & SpeechMatrix$^\ddagger$ & 1451 & 3.36  & 25.65 & 2.50 & 2.87 \\
         \midrule
         \multirow{2}{*}{\textit{\textbf{Baseline}}} & {\lm} & 1500 & \underline{\textbf{3.93}} & \underline{\textbf{28.86}} & 2.53 & 2.91 \\
         & {\gm} & 1000 & 3.42 & 27.84 & 2.49 & 2.86 \\
         \midrule
         \textit{\textbf{Our Method}} & {\sva} & 750 & 3.73 & 28.69 & \underline{\textbf{2.55}} & \underline{\textbf{2.94}} \\
         \midrule
         & \multicolumn{2}{l}{p-value w.r.t {\lm}} & 0.1638 & 0.1938 & 0.0659 & 0.0610 \\
         & \multicolumn{2}{l}{p-value w.r.t {\gm}} & 0.0939 & 0.0050\textsuperscript{*} & 0.0020\textsuperscript{*} & 0.0010\textsuperscript{*} \\
         \midrule
         \midrule
         & \multicolumn{1}{l}{\colorbox{lightgray}{\textbf{German-to-English}}} \\
         \multirow{2}{*}{\textit{\textbf{State-of-the-art}}} & SpeechMatrix$^\dagger$  & 1456 & 8.3  & - & - & -\\
         & SpeechMatrix$^\ddagger$  & 1456 & \underline{12.18}  & 35.08 & \underline{3.16} & 3.51 \\
         \midrule
         \multirow{2}{*}{\textit{\textbf{Baseline}}} & {\lm} & 1250 & 11.52 & 37.88 & 3.09 & 3.46 \\
         & {\gm} & 750 & \textbf{11.55} &  38.24 & 3.12 & 3.49 \\
         \midrule
         \textit{\textbf{Our Method}} & {\sva} & 1000 & 11.39 & \underline{\textbf{38.34}} & \underline{\textbf{3.16}} &  \underline{\textbf{3.53}} \\         
         \midrule
         & \multicolumn{2}{l}{p-value w.r.t {\lm}} & 0.2468 & 0.0829 & 0.0010\textsuperscript{*} & 0.0010\textsuperscript{*} \\
         & \multicolumn{2}{l}{p-value w.r.t {\gm}} & 0.2907 & 0.2118 & 0.0160\textsuperscript{*} & 0.0120\textsuperscript{*} \\
         \bottomrule
    \end{tabular}
    \caption{Results for En-to-De and De-to-En on FLEURS test sets. \textbf{Bold} means better than {\sm} baselines. \underline{Underline} means the best overall.
    $^\dagger$Results from \citet{duquenne-etal-2023-speechmatrix}.
    $^\ddagger$Models trained by ourselves.
    \textsuperscript{*}p-value $<0.05$.
    Results show that {\sva} models perform better than baselines under most metrics in both directions.
    }
    \label{tab:fleurs_result}
\vspace{-10pt}
\end{table*}

\section{Evaluation Results on FLEURS}
\label{app:fleurs_result}
We provide evaluation results on the FLEURS test set in Table \ref{tab:fleurs_result}.
Similar to Section \ref{sec:results}, our results match or outperform SpeechMatrix results.
For both En-to-De and De-to-En, {\sva} and {\gm} achieve comparable performance when using the transcription-based metrics ASR-BLEU and ASR-chrF2++.
Their performance is especially close on De-to-En.
However, {\sva} is significantly better than {\gm} when using the BLASER 2.0 metrics, achieving an improvement of 0.06 and 0.04 referenced BLASER 2.0 scores on En-to-De and De-to-En, respectively.

For En-to-De, {\sva} achieves comparable performance with {\lm} on all metrics.
For De-to-En, {\sva} is significantly better than {\lm} when using BLASER 2.0 metrics.

Overall, we can show that {\sva} performs better than both {\lm} and {\gm}.

\section{Procedure of Manual Alignments}
\label{app:gold_alignment}
The manual alignment procedure is as follows:
\begin{enumerate}
    \item We apply Whisper \cite{Radford-2023-whisper} to transcribe the German and English speech documents;
    \item We manually select the corresponding words for each speech segment from the obtained transcriptions;
    \item We used Google Translate to translate the German transcriptions into English, as the author does not speak German;
    \item We align the German and English segments based on the corresponding English transcriptions and translations.
\end{enumerate}
Although this process depends on models such as Whisper and Google Translate, we argue that they should perform extremely well on German and English and should produce almost perfect transcriptions and translations.

\section{Evaluation of System Alignments using the Manual Alignments}
\label{app:align_eval_with_gold}

We use the same alignment evaluation method as Section \ref{sec:sv_similar_to_sm}, but we use the manual alignments as the reference.
There are 144 raw {\sva} alignments, and we choose the same number of alignments from {\gm} and {\lm} in descending order of margin-scores.
The Recall and Precision of raw {\sva}, {\lm}, and {\gm} alignments are presented in Table \ref{tab:sv_lm_gm_vs_gold}.

The three methods have similar Lax Precisions, while that of {\lm} and {\gm} are slightly higher.
{\sva} has the highest recall values than both the {\sm} baselines.
Among the three methods, {\lm} has the worst performance in general.
This follows Figure~\ref{fig:local_mining} that both {\sva} and {\gm} have good performance but {\lm} does not perform well.

\begin{table}[h]
    \centering
    \small
    \begin{tabular}{lcccc}
        \toprule
         & \multicolumn{2}{c}{\textbf{Precision}} & \multicolumn{2}{c}{\textbf{Recall}} \\
         \cmidrule{2-3} \cmidrule{4-5}
         & \textit{Strict} & \textit{Lax} & \textit{Strict} & \textit{Lax} \\
         \midrule
        raw {\lm} & 0.139 & \textbf{0.993} & 0.147 & 0.676 \\
        {\gm} & 0.188 & \textbf{0.993} & 0.199 & 0.868 \\
        raw {\sva} & \textbf{0.597} & 0.979 & \textbf{0.632} & \textbf{0.978} \\
        \bottomrule
    \end{tabular}
    \caption{Precision and Recall for raw {\sva}, {\gm} and raw {\lm} alignments when manual alignments are used as the reference.}
    \label{tab:sv_lm_gm_vs_gold}
\vspace{-10pt}
\end{table}

\begin{table*}[h]
    \centering
    \footnotesize
    \begin{tabular}{lccc}
        \toprule
        \multicolumn{1}{c}{\textbf{Stage}} & \textbf{\# En Segments} & \textbf{\# De Segments} & \textbf{\# Alignments} \\
        \midrule
        Segmentation & 4,113,319 & 3,056,797 & - \\
        \midrule
        Detection of identical untranslated segments  & 47,008 & 47,008 & - \\
        \midrule
        Segment concatenation & 19,331,307 & 13,012,502 & - \\
        \midrule
        Vecalign & - & - & 2,597,796 \\
        \midrule
        After removing low-quality alignments & - & - & 1,968,141 \\
        \midrule
        After removing untranslated alignments & - & - & 1,962,032 \\
        \midrule
        Alignment concatenation & - & - &  3,661,860 \\
        \midrule
        After removing short alignments ($<$ 1 second) & - & - & 2,810,697 \\
        \midrule
        After removing highly-overlapped alignments  & \multirow{2}{*}{-} & \multirow{2}{*}{-} & \multirow{2}{*}{851,446} \\
        ({\mo} $<$ 0.8 and duration $<$ 2 seconds) & \\
        \midrule
        Choose the best 750 hours & - & - & 317,293 \\
        \bottomrule
    \end{tabular}
    \caption{Number of segments or alignments at each stage.}
    \label{tab:stats_for_intermediate}
\end{table*}

\section{Statistics for Intermediate Procedures}
As our proposed alignment pipeline consists of several intermediate steps, we report numbers of segments or alignments in Table \ref{tab:stats_for_intermediate}.
We use English-to-German alignment as an example.

\end{document}